\documentclass{article}

\usepackage[preprint]{corl_2026}

\usepackage[T1]{fontenc}        
\usepackage[utf8]{inputenc}     


\usepackage{graphicx}
\usepackage{amsmath, amssymb}
\usepackage{booktabs}
\usepackage{multirow}
\usepackage{xcolor}
\usepackage{colortbl}
\usepackage{caption}
\usepackage{makecell}
\usepackage{graphicx}
\usepackage{tabularx}
\usepackage{wrapfig}
\usepackage{subcaption} 
\usepackage{amsmath}

\DeclareMathOperator*{\argmin}{argmin}

\newcommand{\Clift}{\mathcal{C}_{\mathrm{lift}}}

\newcommand{\Cadm}{\mathcal{C}_{\mathrm{adm}}}
\newcommand{\phat}{\hat{\mathbf{p}}}
\newcommand{\Dobs}{D}
\newcommand{\fmodel}{f_{\theta}}
\newcommand{\arc}{\mathrm{arc}}
\newcommand{\Up}[1]{{\small ($+#1\,\%$)}}  
\newcommand{\proj}{\pi}
\newcommand{\projinv}{\pi^{-1}}

\title{PathLift: Depth-Conditioned Metric Lifting\\for Visual-Navigation Priors}

\author{
  Gwanjun Shin$^{1}$ \quad Hyeongseok Suh$^{2}$ \quad Hongmo Jung$^{2}$ \quad Jongwoo Lim$^{2\dagger}$\\
  $^{1}$Interdisciplinary Program in Artificial Intelligence, Seoul National University\\
  $^{2}$Department of Mechanical Engineering, Seoul National University\\
  \texttt{\{shinkansan, hyeongaa, jjhm2910, jongwoo.lim\}@snu.ac.kr}\\
  $^{\dagger}$Corresponding author
}
\hypersetup{
  pdftitle={PathLift: Depth-Conditioned Metric Lifting for Visual-Navigation Priors},
  pdfauthor={Gwanjun Shin, Hyeongseok Suh, Hongmo Jung, Jongwoo Lim}
}

\begin{document}
\maketitle

\begin{abstract}
Generalist visual-navigation policies tell a robot which direction
to go but not how far. A separate lifting step is needed at
deployment to convert their unitless trajectory into a metric path
that a controller can track; existing scalar lifters apply uniform
scaling that misses per-waypoint corrections, while in-policy
approaches that inject monocular depth into the policy inherit the
prediction errors of the depth model near obstacles and at close
range. We present PathLift, a per-waypoint depth-conditioned
adapter that lifts a pre-trained visual-navigation policy by
consuming an explicit, sensor-grade depth source (RGB-D, stereo, or
projected LiDAR) and fitting per-waypoint magnitude and lateral
residuals to a canonicalized prior. PathLift consistently improves closed-loop navigation across
diverse visual-navigation policies in simulation and on a real
wheeled robot under both stereo and LiDAR depth. By turning the
policy's unitless intent into a depth-feasible metric path in a
single feed-forward pass, the adapter lifts real-world success
without test-time optimization.
\end{abstract}


\section{Introduction}
\label{sec:intro}

Visual-navigation policies such as GNM, ViNT, NoMaD, FlowNav, and
NaviBridger~\citep{shah2023gnm,shah2023vint,sridhar2024nomad,gode2025flownav,ren2025navibridger}
tell a robot which way to go but not how far. To enable cross-embodiment generalization, these models normalize their training trajectories by a robot-specific constant tied to top speed. As a result, they output a geometrically meaningful but unitless trajectory, which we refer to as the \emph{visual-navigation prior}. Before a controller can track this prior at deployment time, a separate metric lifting step is required.

Three existing approaches illustrate the gap.
\emph{Scalar lifting} applies one scale uniformly across all
waypoints, reducing the prior's per-waypoint geometry to a single
scale factor; it over- or under-shoots when the effective scale
varies along the trajectory, and it misses the lateral corrections
that depth observation makes possible.
\emph{In-policy monocular depth}~\citep{gode2025flownav,cai2025navdp}
injects a depth pathway into the policy itself, which couples the
geometric pathway to a specific policy; when monocular depth is the
source, the policy also inherits its prediction errors near
obstacles and at close range, precisely where the lift must be
precise.
\emph{Test-time geometric trajectory
optimization}~\citep{ratliff2009chomp,rosmann2012trajectory,zhang2026saferpath}
expresses richer trajectories but requires a map and per-frame
optimization, and consumes the prior only after a lift has
placed it in metric space (Sec.~\ref{sec:planners}).

\begin{figure}[!htbp]
\centering
\includegraphics[width=0.98\textwidth]{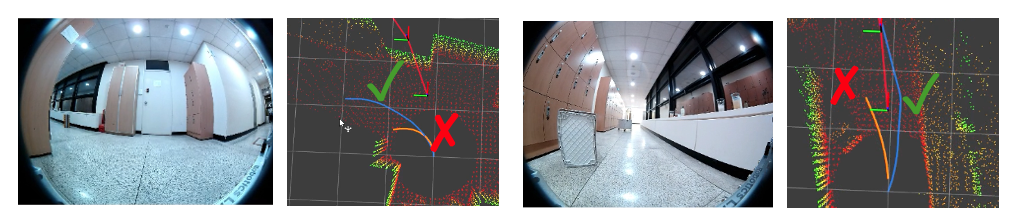}
\caption{\textbf{PathLift in action.} The raw NoMaD output (red)
is the unitless prior. Lifted by a const-scalar baseline (orange,
$\times$), the trajectory drives the robot into structural
obstacles; lifted by PathLift (blue, $\checkmark$), it routes
around them. Each pair shows the fisheye RGB and the corresponding
BEV with occupied cells (black) and depth points (brown).}
\label{fig:teaser}
\end{figure}

PathLift (Fig.~\ref{fig:teaser}) is a per-waypoint depth-conditioned
\emph{adapter} that attaches to any pre-trained visual-navigation
policy. It takes shape from a canonicalized prior and scale from an
explicit, sensor-grade depth source projected into the policy's
camera frame, supervised at training by a structured admissible
family of intent-preserving metric trajectories.

We summarize our contributions:
%
%
\begin{itemize}\setlength\itemsep{0.2em}
    \item \textbf{Depth-feasible metric path in a single pass.}
    PathLift's training target is selected from an
    intent-preserving, depth-feasible subset of candidate
    lifts. The deployed adapter turns each prior sample into
    a depth-feasible metric path in a single forward pass.

    \item \textbf{Prior-agnostic, sensor-agnostic adapter.}
    A single checkpoint lifts both regression-based (GNM, ViNT) and generative (NoMaD, FlowNav) priors without retraining, and accepts stereo, RGB-D, or
    projected LiDAR depth through a single camera-frame interface.

    \item \textbf{Low-latency adapter.}
    PathLift adds $5$\,ms adapter overhead per inference, over $10\times$
    faster than the in-policy lifting baseline (Tab.~\ref{tab:sim_main}).
\end{itemize}

\section{Related Work}
\label{sec:related}

\paragraph{Visual-navigation policies and foundation models.} The
landscape of generalist navigation models is expanding rapidly,
from goal-conditioned image-frame
predictors~\citep{shah2023gnm,shah2023vint,sridhar2024nomad,gode2025flownav,ren2025navibridger,shen2025effonav}
to navigation foundation models trained on
web-scale data~\citep{liu2025citywalker,zhang2025embodied,hirose2025omnivla}
and world-model-based planners~\citep{bar2025nwm}. A common trait
remains: their outputs are shape-correct but unitless trajectories
in an embodiment-agnostic coordinate frame, requiring a separate
lifting step at deployment regardless of how the policy is trained.
Without geometric grounding, these policies collide frequently
when deployed~\citep{guerrier2026can}.
A complementary trend scales these models on web video by
recovering action labels via deep visual SLAM~\citep{teed2021droid, teed2023deep}
or structure-from-motion~\citep{schonberger2016structure,leroy2024grounding, wang2025vggt}, as in
CityWalker~\citep{liu2025citywalker}; the recovered trajectories
expose only shape, with deployment scale supplied by a per-robot
constant.

\paragraph{Lifting approaches.} The common recipe applies a hand-tuned
$v_{\max}/f$ scalar uniformly across waypoints; learned scalar
lifters~\citep{nayak2025metricnet} replace it with a regressed
single scale, still uniformly applied. A second line couples depth
into the policy itself, either through monocular-depth
features~\citep{gode2025flownav,yang2024depth} or through a
dedicated depth-input encoder~\citep{cai2025navdp}. Both choices
tie the depth pathway to a specific policy: monocular features
inherit prediction errors near obstacles and at close range, and
dedicated depth encoders require retraining when the sensor
changes. PathLift instead consumes an explicit, sensor-grade depth
source (stereo, RGB-D, or projected LiDAR) at the adapter
interface, keeping the upstream policy untouched.

\paragraph{Test-time trajectory optimization.} Recent visual-navigation
optimizers~\citep{zhang2026saferpath} solve a smooth-cost problem
against a depth-derived traversability field per frame, following a formulation traceable to classical trajectory
optimization~\citep{ratliff2009chomp,rosmann2012trajectory,schulman2014motion}. These methods
solve a single instance optimally given a map.

PathLift offers a distinct approach: an adapter that couples a pre-trained visual-navigation policy with feasible depth via a feed-forward mapping integrating the prior’s intent, depth-observed free space, and expert navigation behavior in a single pass. Test-time collision optimization trades the first of these signals against the second through soft costs; PathLift instead bounds deviation from the prior as a hard constraint, without optimizing an obstacle cost. This adapter design provides compatibility with diverse pre-trained policies and arbitrary depth sources without retraining.

\section{Method}
\label{sec:method}

\begin{figure}[t]
\centering
\includegraphics[width=0.98\textwidth]{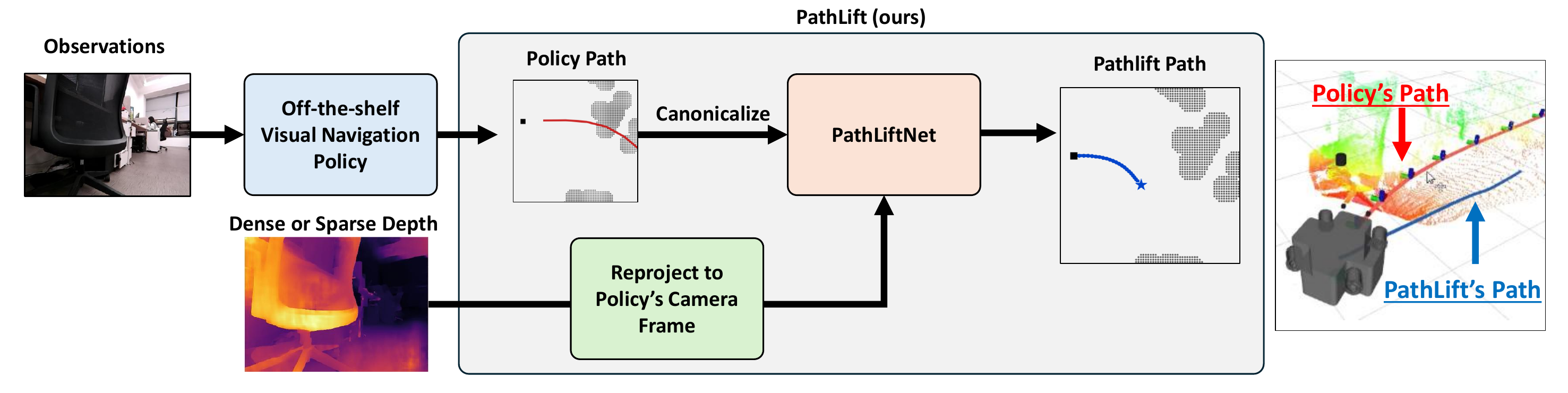}
\caption{\textbf{PathLift inference pipeline.} An off-the-shelf
visual-navigation policy emits a unitless policy path; the
canonicalized prior and a depth observation (RGB-D, stereo, or
projected LiDAR) reprojected into the policy's camera frame feed
PathLiftNet, which outputs a metric trajectory in body coordinates.}
\label{fig:arch}
\end{figure}

\subsection{Problem formulation}
\label{sec:problem}

We formulate metric lifting as structured-target regression. The
visual-navigation prior is canonicalized to a unitless sequence of
$H$ waypoints $\phat\in\mathbb{R}^{H\times 2}$ by origin-aligning,
arc-length-resampling, and rescaling so that $\|\phat_H\|_2=1$,
giving a common input format across policies. The prior's raw scale
follows training convention and is not aligned with the deployment
scene, so metric scale must be recovered from depth.

Given $\phat$, a depth frame $\Dobs$, and the source-camera
projection $\proj$ (intrinsics and the camera-to-body
transform $T_{BC}$), jointly $\mathcal{O}=(\Dobs,\proj)$, the
network predicts a body-frame trajectory $y\in\mathbb{R}^{H\times 2}$
that tracks an expert path $y^{\star}$ within the depth-observed
free space. For each waypoint $k\in\{1,\dots,H\}$:
\begin{equation}
\label{eq:output}
y_k \;=\; r_k\,\phat_k \;+\; \delta_k\,\mathbf{n}_k(\phat),
\end{equation}
where $r_k\in\mathbb{R}_{+}$ scales the prior magnitude at waypoint
$k$ and $\delta_k\in\mathbb{R}$ applies a lateral residual along the
per-waypoint unit normal $\mathbf{n}_k(\phat)$.

The training data provides expert paths $y^{\star}$ but no
ground-truth lifts: an expert path is where the robot traveled,
not a metric realization of the prior's shape. Regressing onto
$y^{\star}$ directly would let the model discard the prior and
leave no way to enforce depth feasibility. At training we therefore
select the supervision target from the \emph{lifted candidate
family} $\Clift(\phat)$, a finite set of candidates spanning radial scale, progress warp, rotation, and
bounded lateral correction (ten sub-families,
App.~\ref{app:tiers}), keeping the depth-feasible candidate closest
to $y^{\star}$. $\Clift$ enters only at training, to define the supervision target.

\subsection{Supervision target within \texorpdfstring{$\Clift$}{Clift}}
\label{sec:supervision}

To ensure the supervision target is physically viable, we restrict $\Clift(\phat)$ to a depth-feasible subset $\Cadm(\phat,\mathcal{O})$ defined by four admissibility predicates:
\begin{equation}
\label{eq:adm}
\Cadm(\phat,\mathcal{O}) =
\bigl\{\,c\in\Clift(\phat) \,:\,
\mathrm{clr}(c)\!\ge\!\tau_\mathrm{clr},\;
\mathrm{pen}(c)\!=\!0,\;
\mathrm{obs}(c)\!\ge\!\tau_\mathrm{obs},\;
\mathrm{tube}(c)\!\le\!\tau_\mathrm{tube}\,\bigr\}.
\end{equation}

These predicates filter candidates based on safety, visibility, and intent preservation:
\begin{itemize}
    \item \textbf{Safety:} We enforce a minimum clearance $\mathrm{clr}(c)\!=\!\min_k\Omega(c_k)$ and a zero penetration count $\mathrm{pen}(c)\!=\!|\{k\!:\!\Omega(c_k)\!<\!\tau_\mathrm{pen}\}|$, where $\Omega$ is the body-frame clearance field. We use $\tau_\mathrm{clr}\!=\!0.30$\,m for a comfort margin and $\tau_\mathrm{pen}\!=\!0.20$\,m as a catastrophic safety floor.
    \item \textbf{Visibility:} The trajectory must stay within the 
    depth-observed region of the BEV. We require 
    $\mathrm{obs}(c) = \tfrac{1}{H}|\{k:c_k \in \mathcal{V}\}| \ge \tau_\mathrm{obs}$, 
    where $\mathcal{V} \subset \mathbb{R}^2$ is the BEV projection of 
    depth-observed free space. Waypoints outside $\mathcal{V}$ enter 
    regions where occupancy is unknown, so their feasibility cannot 
    be certified. We set $\tau_\mathrm{obs} = 0.90$.
    \item \textbf{Intent Preservation:} To prevent excessive deviation 
    from the prior's shape, we bound the average residual from its 
    best-fit pure-radial lift, $\mathrm{tube}(c) = \tfrac{1}{H}\sum_k 
    \|c_k - \alpha_c\hat{p}_k\| \le \tau_\mathrm{tube}$, where 
    $\alpha_c = \langle c,\hat{p}\rangle/\|\hat{p}\|^2$ is the 
    least-squares scalar fit aligning $\hat{p}$ with $c$. Pure radial 
    candidates ($c = R\hat{p}$) give zero residual. Rotation or lateral 
    correction increases it. We use $\tau_\mathrm{tube} = 0.30$\,m,
    the tightest bound that costs little in supervision
    quality (App.~\ref{app:sensitivity}).
\end{itemize}

\begin{figure}[!htbp]
\centering
\includegraphics[width=0.96\textwidth]{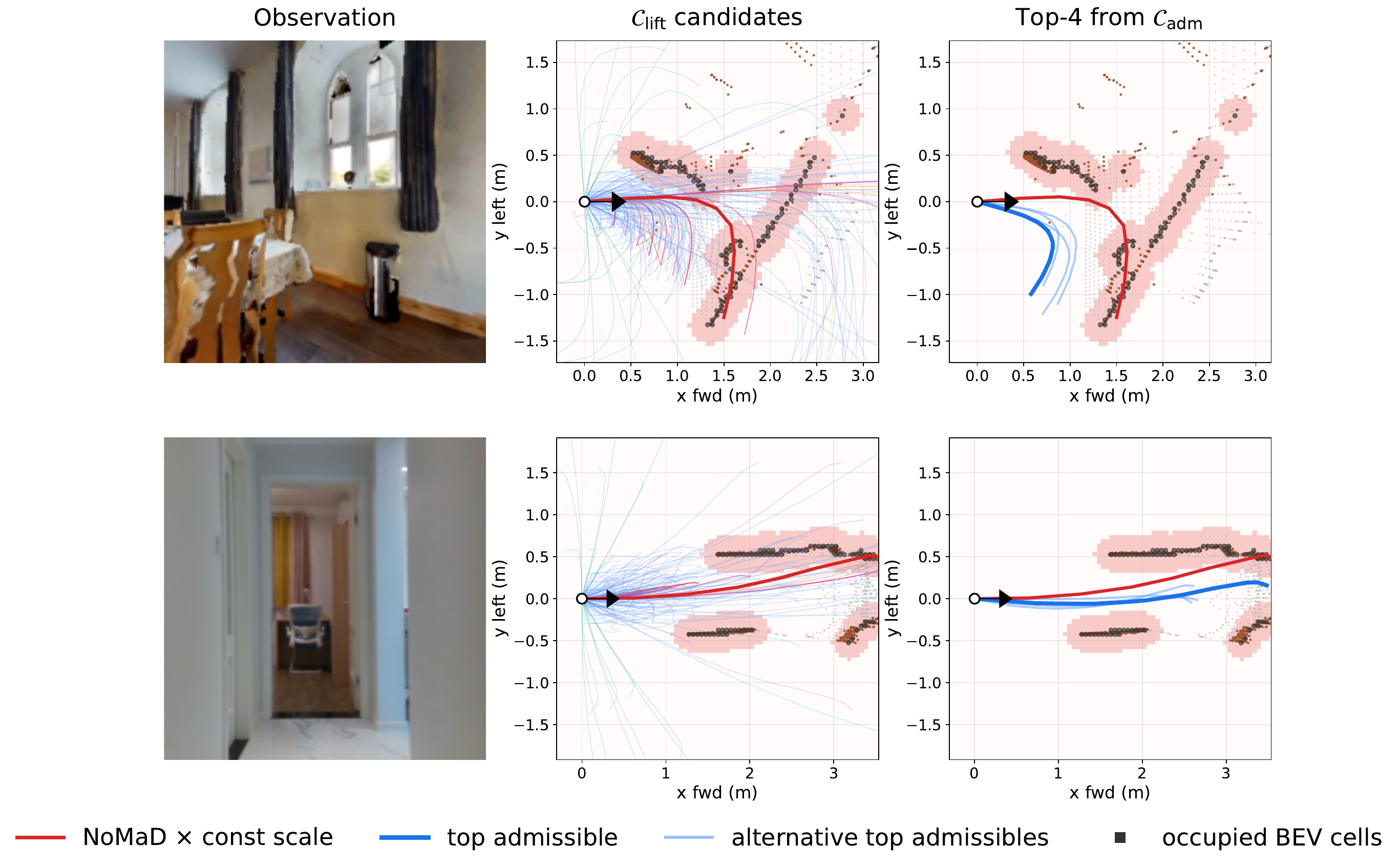}
\caption{Supervision target on Habitat (top) and OpenLORIS
(bottom). Col.~1: camera RGB. Col.~2: all candidates in
$\Clift(\phat)$ from the NoMaD prior (faded). Col.~3: four
lowest-ADE admissibles; supervision target in solid blue. The
NoMaD $\times$ const scale (red) and occupied BEV cells (black)
overlaid in both BEV columns.}
\label{fig:qualitative}
\end{figure}

At training time, the admissible candidate closest to ground truth
$y^\star$ in average displacement error (ADE) is the supervision
target,
\begin{equation}
\label{eq:pick}
c^\star \;=\; \argmin_{c\in\Cadm(\phat,\mathcal{O})}\, \mathrm{ADE}(c, y^\star).
\end{equation}
Ground truth $y^\star$ enters only at training, to select $c^\star$
from $\Cadm$. This supervision combines three signals in a single
forward pass: the prior's intent (via $\Clift$), the depth-observed
free space (via $\Cadm$), and expert navigation behavior (via the
closest admissible candidate to $y^\star$).
Fig.~\ref{fig:qualitative} illustrates this construction. The candidate family and admissibility predicates are shared 
across all training data regardless of source policy or 
camera model.

\paragraph{Scalar lifts are suboptimal.}
Empirical analysis justifies our selection-based target
construction over simple scalar lifting. In indoor environments
(Habitat~\citep{savva2019habitat}, OpenLORIS~\citep{shi2020we}), pure-radial candidates, representing
the single-scalar regime, account for only $3.1$--$3.7\%$ of
$c^\star$ selections. The remaining $>96\%$ of optimal lifts
require progress-dependent scaling, rotation, or lateral
corrections. Outdoor driving (CODa~\citep{zhang2024toward}) is an exception
($27$--$34\%$) due to predominantly straight paths, but the
indoor results strongly motivate a per-waypoint output over a
single global scalar. Per sub-family admissibility pass rates and
the resulting subset composition are reported in
App.~\ref{app:tiers}.

\subsection{Architecture}
\label{sec:arch}

The PathLift adapter is a feed-forward network
$\fmodel:(\phat,\mathcal{O})\to y$ that consumes the canonicalized
prior and an explicit depth observation, mapping them directly to
a metric body-frame trajectory (Fig.~\ref{fig:arch}).

\paragraph{Tokens.} We discretize the depth frame into $256$ patch tokens of dimension $16$. Each token combines three geometric signals: (i) seven depth statistics within the patch (the median, the $p10/p25/p75/p90$ quantiles, the variance, and the valid-pixel fraction), (ii) the six coordinates of a Pl\"ucker ray~\citep{sitzmann2021light} for the patch's central pixel in body frame, and (iii) the three coordinates of a body-frame anchor at the patch's median-depth point. Prior tokens of dimension $12$ encode each waypoint's canonicalized position, progress fraction, and Pl\"ucker line. The unprojection $\projinv$ supports pinhole, double-sphere fisheye~\citep{usenko2018double}, and equirectangular models; once tokenized, the network consumes them uniformly. Any metric depth source (RGB-D or LiDAR) is pre-aligned via re-projection through $\proj$ into the policy's camera frame, decoupling the adapter from the sensor.

\paragraph{Network.} A 4-layer scene encoder $E_{\mathrm{scene}}$
and a 2-layer prior encoder $E_{\mathrm{prior}}$ process the patch
and prior tokens with self-attention. An anchor MLP maps the prior
tokens to $H$ anchor queries, which a cross-attention block primes
with depth context. Six decoder
blocks~\citep{shi2022motion} iteratively refine the queries via
alternating self- and cross-attention to $E_{\mathrm{scene}}$ and
$E_{\mathrm{prior}}$. Each refined query emits a magnitude $r_k$
and a lateral residual $\delta_k$; the trajectory follows
Eq.~\eqref{eq:output}. The full network has $4.3$\,M parameters.

\paragraph{Training.}
Training combines an MSE term to $c^\star$ (Eq.~\eqref{eq:pick})
with clearance, prior-coupling, smoothness, and deep-supervision
losses~\citep{lee2015deeply}. A single adapter is trained on
Habitat pinhole and fisheye renders, CODa projected LiDAR and
OpenLORIS RGB-D, with priors drawn uniformly from GNM, ViNT and
NoMaD (App.~\ref{app:loss}).

\section{Experiments and Results}
\label{sec:experiments}

We evaluate PathLift in closed-loop benchmarks in simulation and on
a real wheeled robot, paired with diverse visual-navigation priors.

\subsection{Simulation Experiments}
\label{sec:sim_benchmark}
\subsubsection{Experimental Setup}

\paragraph{Setup and Baselines.}
We benchmark PathLift on the HM3D dataset
\citep{ramakrishnan2021hm3d} within the Habitat-sim environment
\citep{savva2019habitat}. The benchmark consists of 2,000 trials
across 50 unseen scenes (20 trajectories per scene, 2 repetitions,
mean length of 27.1 nodes). Navigation is executed in a closed-loop
manner via waypoint tracking. We terminate a trial upon the first
collision or if the robot fails to reach the goal within 500 steps.
Detailed sensor and robot specifications are provided in
App.~\ref{app:sim_robot}.
We evaluate four state-of-the-art image-goal policies with several baselines:
\begin{itemize}
    \item \textbf{Visual-navigation policies:} We include regression-based
    feed-forward models, GNM \citep{shah2023gnm} and ViNT
    \citep{shah2023vint}, alongside generative policies based on diffusion
    and flow-matching, NoMaD \citep{sridhar2024nomad} and FlowNav
    \citep{gode2025flownav}. All four policies output waypoint trajectories
    in normalized (unitless) form. All policies are fine-tuned on HM3D
    training scenes disjoint from the evaluation set
    (App.~\ref{app:finetuning}).

    \item \textbf{Adapters:} We compare PathLift against 
    MetricNet/MetricNav~\citep{nayak2025metricnet}.
    Since MetricNav injects guidance during the denoising steps, it is
    directly compatible only with generative priors. To isolate the impact
    of the lifting algorithm from the depth source, we report
    MetricNav$^{*}$, which builds the baseline's TSDF map from the same
    ground-truth depth used by PathLift, leaving MetricNet unchanged.
    MetricNav is trained separately for each prior, whereas the same PathLift
    adapter attaches to all four priors without retraining
    (App.~\ref{app:metricnav_star}).
\end{itemize}

\paragraph{Metrics.}
We report Success Rate (SR), Success weighted by Path
Length~\citep{anderson2018evaluation} (SPL), Topological-Map Reached (TR),
and inference time. TR denotes the fraction of topological nodes
traversed before trial termination, providing a finer-grained progress
signal than the binary SR. Inference time is measured on an NVIDIA
RTX 4090 for the prior and adapter combined.

\begin{table*}[t]
\centering
\footnotesize
\setlength{\tabcolsep}{3pt}
\setlength{\abovecaptionskip}{0pt}
\setlength{\belowcaptionskip}{12pt}
\caption{Closed-loop simulation benchmark on 50 HM3D scenes
(2{,}000 trials per row).}
\vspace{1.0em}
\label{tab:sim_main}

\begin{tabular}{llrrrrr}
\toprule
\textbf{Base prior} & \textbf{Adapter} &
\textbf{SR \% ($\Delta$)} &
\textbf{SPL \% ($\Delta$)} &
\textbf{TR \% ($\Delta$)} &
\makecell[r]{\textbf{Adapter}\\\textbf{(ms)}} &
\makecell[r]{\textbf{Total}\\\textbf{(ms)}} \\
\midrule

GNM
& --
& 17.40
& 17.14
& 39.03
& ---
& \textbf{2.6} \\

\rowcolor{gray!12}
GNM
& PathLift
& \textbf{25.60} \Up{47.1}
& \textbf{24.54} \Up{43.2}
& \textbf{49.39} \Up{26.5}
& \textbf{5.3}
& \textbf{8.2} \\

\midrule

ViNT
& --
& 25.00
& 24.57
& 48.34
& ---
& \textbf{5.9} \\

\rowcolor{gray!12}
ViNT
& PathLift
& \textbf{33.70} \Up{34.8}
& \textbf{32.61} \Up{32.7}
& \textbf{55.96} \Up{15.8}
& \textbf{5.2}
& \textbf{11.1} \\

\midrule

NoMaD
& --
& 17.45
& 17.25
& 44.00
& ---
& \textbf{24.9} \\

NoMaD
& MetricNet
& 28.95 \Up{65.9}
& 28.05 \Up{62.6}
& 53.42 \Up{21.4}
& 10.7
& 35.6 \\

NoMaD
& MetricNav
& 25.10 \Up{43.8}
& 24.42 \Up{41.6}
& 50.53 \Up{14.8}
& 76.1
& 101.8 \\

NoMaD
& MetricNav$^{*}$
& 29.20 \Up{67.3}
& 28.37 \Up{64.5}
& \textbf{53.95} \Up{22.6}
& 103.7
& 129.7 \\

\rowcolor{gray!12}
NoMaD
& PathLift
& \textbf{29.80} \Up{70.8}
& \textbf{28.53} \Up{65.4}
& 53.04 \Up{20.5}
& \textbf{5.3}
& \textbf{31.3} \\

\midrule

FlowNav
& --
& 16.00
& 15.94
& 42.27
& ---
& \textbf{28.3} \\

FlowNav
& MetricNet
& 27.05 \Up{69.1}
& 26.12 \Up{63.9}
& 51.82 \Up{22.6}
& 10.8
& 37.6 \\

FlowNav
& MetricNav
& 29.75 \Up{85.9}
& 29.10 \Up{82.6}
& 53.89 \Up{27.5}
& 76.6
& 104.8 \\

FlowNav
& MetricNav$^{*}$
& \textbf{31.60} \Up{97.5}
& \textbf{31.04} \Up{94.7}
& \textbf{54.69} \Up{29.4}
& 103.4
& 130.9 \\

\rowcolor{gray!12}
FlowNav
& PathLift
& 28.45 \Up{77.8}
& 27.45 \Up{72.2}
& 52.01 \Up{23.0}
& \textbf{5.3}
& \textbf{32.1} \\

\bottomrule
\end{tabular}

\vspace{0.35em}
\begin{minipage}{0.98\textwidth}
\scriptsize
\textit{Note:} ``--'' denotes the bare prior with a fixed-scale lift.
Adapter runtime excludes the base-prior forward pass; total runtime is timed
end to end, so the columns need not sum.
\end{minipage}

\end{table*}

\subsubsection{Results and Analysis}

\paragraph{Performance consistency across visual-navigation priors.}
Table~\ref{tab:sim_main} summarizes the closed-loop benchmark results for the 14 combinations of base priors and adapters. PathLift improves all
four base priors without retraining, recovering metric scale from their waypoint outputs, with relative SR gains of
$+34.8\%$ (ViNT) to $+77.8\%$ (FlowNav) and an average TR gain of
$9.19$\,pp. The improvement also generalizes to FlowNav, which was not used during PathLift training.

\paragraph{Comparison with MetricNet and MetricNav.}
MetricNet provides a direct scalar-lifting baseline on the generative
priors. PathLift improves SR over MetricNet by $0.85$\,pp on NoMaD
($29.80\%$ vs.\ $28.95\%$) and $1.40$\,pp on FlowNav
($28.45\%$ vs.\ $27.05\%$), while using roughly half the adapter
runtime ($5.3$ vs.\ $10.7$--$10.8$\,ms).

Is geometric lifting alone enough to match an adapter that also
selects among generated samples and steers them toward the goal?
PathLift matches MetricNav$^{*}$ on NoMaD (SR $29.80\%$ vs.\
$29.20\%$) and trails by $3.15$\,pp on FlowNav. MetricNav$^{*}$
operates inside the denoising loop, drawing multiple trajectory
samples per step and scoring them against a goal-informed TSDF
cost. PathLift instead lifts a single sample per inference,
producing a depth-feasible path without sample selection or goal
scoring; even on a sample not best aligned with the goal, the
feasible lift yields safe progress that the closed loop can build
on. PathLift's adapter overhead is prior-independent and over an
order of magnitude smaller ($5$ vs $76$--$104$\,ms;
Tab.~\ref{tab:sim_main}), and the same adapter attaches to both
regression-based (GNM, ViNT) and generative (NoMaD, FlowNav) priors.

\subsection{Real-Robot Deployment}
\label{sec:real_robot}

\subsubsection{Image-Goal Navigation}
\label{sec:real_imagegoal}

\begin{figure}[t]
  \centering

  \subfloat[]{%
    \resizebox{0.58\linewidth}{4cm}{%
      \includegraphics{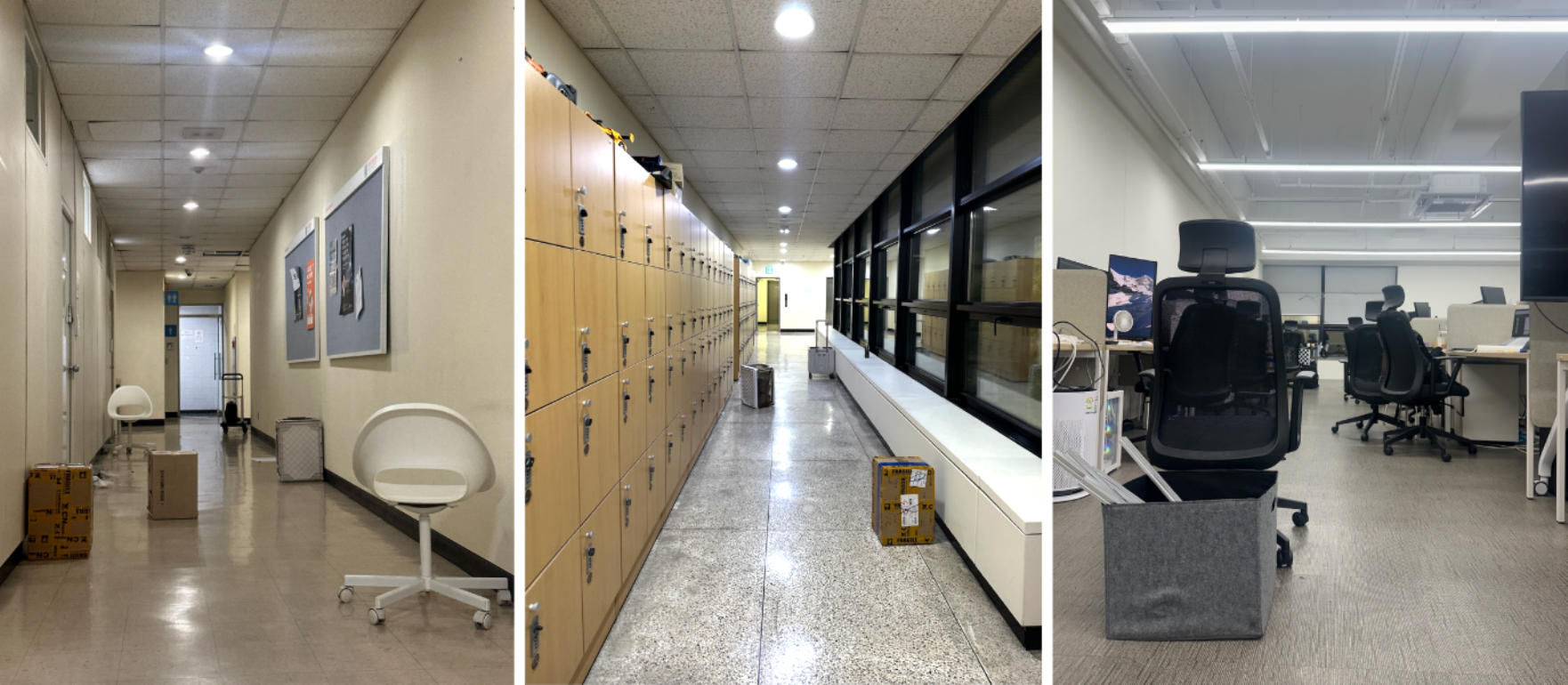}
    }
    \label{fig:real_robot_a}
  }
  \hfill
  \subfloat[]{%
    \resizebox{0.36\linewidth}{4cm}{%
      \includegraphics{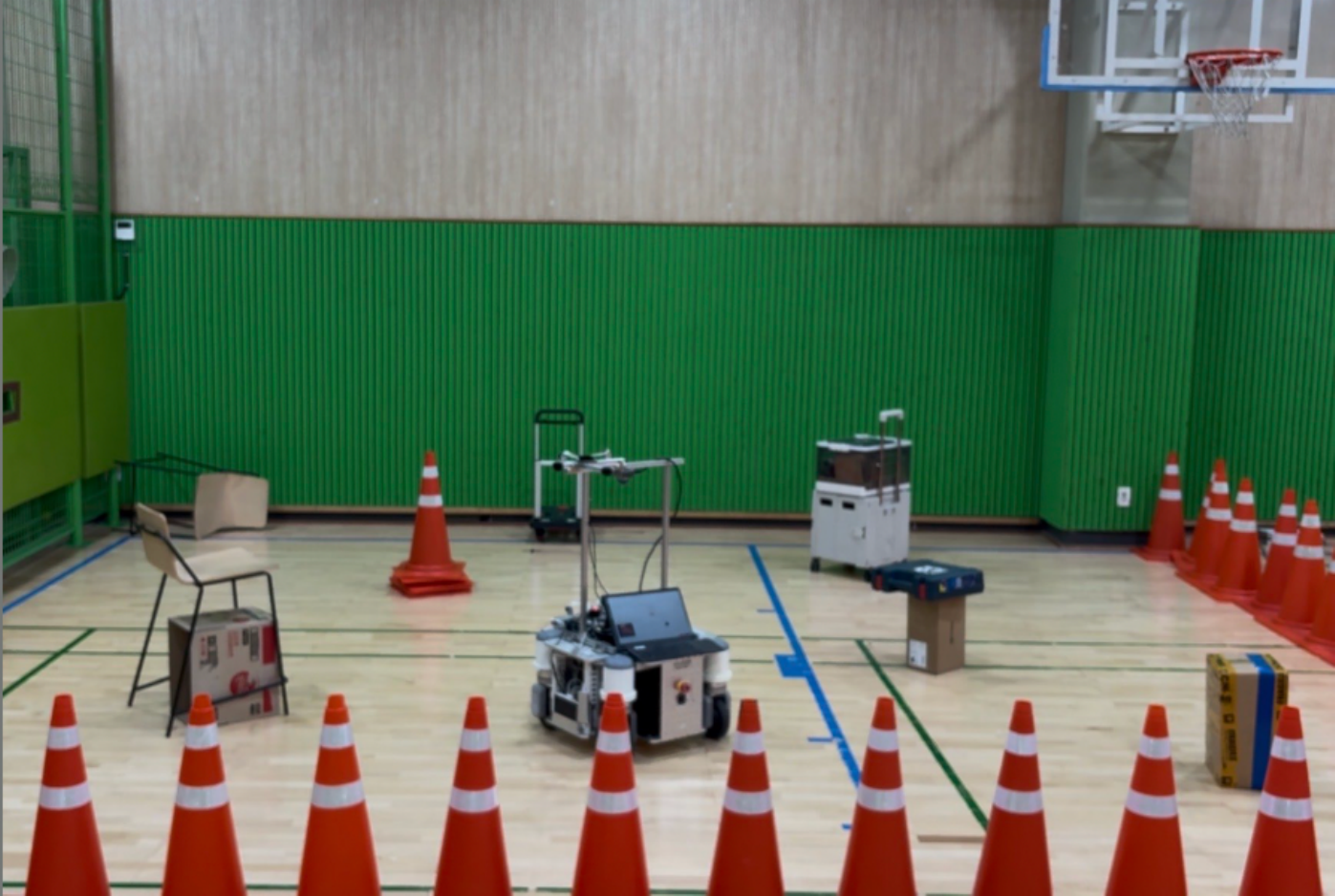}
    }
    \label{fig:real_robot_b}
  }

\caption{Real-robot experimental environments.
  (a)~Image-goal navigation in E1/E2 ($20$\,m corridors)
  and E3 ($10$\,m corridor), shown left to right
  (Sec.~\ref{sec:real_imagegoal}).
  (b)~Goal-free exploration in a $12\!\times\!10$\,m indoor area
  without a pre-built map (Sec.~\ref{sec:real_exploration}).}
  \label{fig:real_robot}
  \vspace{-0.8em}
\end{figure}

\paragraph{Setup and Baselines.}
We deploy PathLift on an indoor wheeled robot with a forward-facing
stereo RGB camera and a LiDAR. The depth
pathway accepts either stereo depth (SD) from a real-time stereo
model~\citep{wen2025fast} or LiDAR rays re-projected into the camera frame.
In each of three different indoor corridor environments (Fig.~\ref{fig:real_robot_a}; two $20$\,m, one 
$10$\,m), the topological map is pre-built without obstacles, and obstacles are inserted before each trial
to create layouts the prior cannot anticipate. A trial succeeds only when
the robot terminates within $1$\,m of the final node; \#Coll counts
every contact event. Platform and inference details are in
Appendix~\ref{app:real_deployment}. We evaluate three policies from Sec.~\ref{sec:sim_benchmark} (GNM,
ViNT, NoMaD) under three adapter conditions: the bare prior scaled by a fixed scale,
MetricNet~\citep{nayak2025metricnet}, and PathLift.
MetricNet is the scalar-scaling front stage of the MetricNav
pipeline (Sec.~\ref{sec:sim_benchmark}); for runtime we omit
MetricNav's downstream goal- and collision-cost refinement and apply
MetricNet only to NoMaD. PathLift uses the same
weights as Sec.~\ref{sec:sim_benchmark}, with no adaptation to the
robot, the scene, or the priors at test time.

\begin{table*}[t]
\centering\footnotesize
\setlength{\tabcolsep}{5pt}
\renewcommand{\arraystretch}{0.7}
\caption{Per-policy image-goal results across three environments (E1, E2, E3); $5$ trials per environment cell, $15$ trials per row in total. SR in \%, \#Coll is mean collisions per trial.}
\label{tab:real_imagegoal_perpolicy}
\begin{tabular}{l ccc ccc ccc c}
\toprule
& \multicolumn{2}{c}{\textbf{E1}}
& \multicolumn{2}{c}{\textbf{E2}}
& \multicolumn{2}{c}{\textbf{E3}}
& \multicolumn{1}{c}{\textbf{Total}} \\
\cmidrule(lr){2-3}\cmidrule(lr){4-5}\cmidrule(lr){6-7}\cmidrule(lr){8-8}
\textbf{Method}
& SR\,$\uparrow$ & \#Coll\,$\downarrow$
& SR\,$\uparrow$ & \#Coll\,$\downarrow$
& SR\,$\uparrow$ & \#Coll\,$\downarrow$
& SR\,$\uparrow$ \\
\midrule
GNM                              & $20$           & $1.6$             & $80$ & $2.2$ & $40$ &
$2.0$ & $46.67$ \\
\rowcolor{gray!12}
GNM $+$\,PathLift (SD)           & $\mathbf{80}$  & $\mathbf{1.0}$   & $\mathbf{100}$ & $\mathbf{1.0}$ & $\mathbf{80}$ &
$\mathbf{1.0}$ & $\mathbf{86.67}$ \\
\midrule
ViNT                             & $80$           & $1.4$             & $60$ & $2.2$ & $60$ &
$1.6$ & $66.67$ \\
\rowcolor{gray!12}
ViNT $+$\,PathLift (SD)          & $\mathbf{100}$ & $\mathbf{0.4}$   & $\mathbf{100}$ & $\mathbf{0.6}$ & $\mathbf{80}$ &
$\mathbf{0.8}$ & $\mathbf{93.33}$ \\
\midrule
NoMaD                            & $40$           & $1.6$      & $20$ & $1.2$ & $40$ &
$1.4$ & $33.33$ \\
NoMaD $+$\,MetricNet~\citep{nayak2025metricnet}
                                 & $40$           & $1.4$     & $40$ & $1.2$ & $80$ &
$1.2$ & $53.33$ \\
\rowcolor{gray!12}
NoMaD $+$\,PathLift (SD)         & $\mathbf{80}$  & $\mathbf{0.8}$  & $\mathbf{80}$ & $\mathbf{0.8}$ & $\mathbf{80}$ &
$\mathbf{1.0}$ & $\mathbf{80}$ \\
\bottomrule
\end{tabular}
\vspace{-0.5em}
\end{table*}

\begin{wrapfigure}{r}{0.45\linewidth}
\vspace{-1.2em}
\centering
\includegraphics[width=\linewidth]{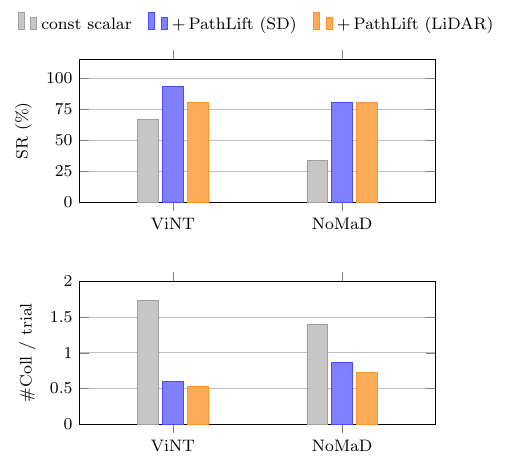}
\caption{Results on different depth sources: success rate (top) and
  collisions per trial (bottom).}
\label{fig:real_robot_FFST_lidar_comparison}
\vspace{-3.5em}
\end{wrapfigure}

\paragraph{Real-world consistency across priors.}
Across the three evaluated policies
(Tab.~\ref{tab:real_imagegoal_perpolicy}),
PathLift with SD achieves a mean $86.7\%$ SR at $0.82$ collisions
per trial, above the bare-prior baselines
($48.9\%$ SR / $1.69$ collisions).
Success rates improve for every policy using the same
PathLift adapter without any per-prior adaptation.
In contrast, MetricNet on NoMaD improves SR by $20$\,pp, below
PathLift's $46.7$\,pp gain on the same prior,
consistent with our claim that
per-waypoint lifting captures scale variation a uniform scalar
cannot.

\paragraph{Consistency across depth sources.}
PathLift yields comparable performance regardless of the depth
source (Fig.~\ref{fig:real_robot_FFST_lidar_comparison}), producing
comparable collision rates on ViNT ($0.60$ SD vs.\ $0.53$ LiDAR) and
NoMaD ($0.87$ SD vs.\ $0.73$ LiDAR) while maintaining an aggregated
SR within $\sim\!7$\,pp between SD ($86.7\%$) and LiDAR ($80.0\%$,
$n{=}30$ each). These results confirm that our depth interface is
decoupled from the sensor hardware, unifying the two sources via
camera-frame re-projection (App.~\ref{app:real_platform}).

\subsubsection{Goal-Free Exploration}
\label{sec:real_exploration}

\paragraph{Setup.}
To probe PathLift's contribution beyond image-goal navigation, we
also evaluate goal-free exploration on the same robot. In the
$12\!\times\!10$\,m indoor area of Fig.~\ref{fig:real_robot_b},
a robot runs in its no-goal exploration mode from a fixed start pose;
each trial ends at the first collision, and the protocol is repeated
$10$ times per method. We compare the bare NoMaD prior against
NoMaD$+$PathLift (SD, same weights as Sec.~\ref{sec:real_imagegoal}).

\begin{table}[t]
  \centering
  \footnotesize
  \setlength{\tabcolsep}{5pt}
  \caption{Real-robot goal-free exploration; NoMaD in no-goal mode,
    $10$ trials per method, run until first collision. Distances and
    durations are measured from the start pose to the contact event.}
  \label{tab:real_exploration}
  \vspace{0.5em}
  \begin{tabular}{lrrrr}
    \toprule
    \textbf{Method} & \textbf{mean dist.\,$\uparrow$ (m)}
      & \textbf{max dist.\,$\uparrow$ (m)}
      & \textbf{mean dur.\,$\uparrow$ (s)}
      & \textbf{max dur.\,$\uparrow$ (s)} \\
    \midrule
    NoMaD                       & $7.86$           & $13.38$          & $29.79$          & $47.57$           \\
    \rowcolor{gray!12}
    \textbf{NoMaD $+$ PathLift} & $\mathbf{11.67}$ & $\mathbf{22.50}$ & $\mathbf{70.66}$ & $\mathbf{111.87}$ \\
    \bottomrule
  \end{tabular}
\end{table}

\paragraph{Longer collision-free exploration.}
PathLift extends the mean trajectory before first collision from
$7.86$ to $11.67$\,m ($+48\%$) and the maximum from $13.38$ to
$22.50$\,m ($+68\%$); mean and maximum durations rise from
$29.79$/$47.57$\,s to $70.66$/$111.87$\,s, more than doubling the time
before contact (Tab.~\ref{tab:real_exploration}).

\begin{table}[!htbp]
\centering\small
\setlength{\tabcolsep}{6pt}
\caption{Ablations on NoMaD over an $800$-trial HM3D subset. Each
variant removes one component from PathLift's default configuration.
Full results in Appendix~\ref{app:ablations}.}
\vspace{0.5em}
\label{tab:ablations}
\begin{tabular}{lc}
\toprule
\textbf{Variant} & \textbf{SR (\%)} \\
\midrule
\rowcolor{gray!12}
PathLift                                              & $\mathbf{29.62}$ \\
$-$\,depth input                          & $6.50$ \\
$-$\,Pl\"ucker tokenizer (CNN encoder)                & $14.25$ \\
$-$\,Radial-only (per-waypoint)       & $27.62$ \\
$-$\,$\Clift$ supervision (direct $y^\star$ regression) & $14.88$ \\
\midrule
Bare NoMaD (fixed-scale lift)                 & $19.88$ \\
\bottomrule
\end{tabular}
\vspace{-1.0em}
\end{table}

\subsection{Ablations}
\label{sec:exp_ablation}

We ablate four PathLift design choices on the NoMaD prior over an
$800$-trial HM3D subset (bare NoMaD: $19.88\%$ SR;
Tab.~\ref{tab:ablations}).

\emph{Does PathLift use depth meaningfully?}
Removing the depth input reduces SR to $6.50\%$, below the bare-prior
baseline. An open-loop sweep over the fraction of invalid depth pixels
shows little change until the input becomes entirely invalid
(App.~\ref{app:ablations}).

\emph{Does Pl\"ucker tokenization beat raw-depth CNN?}
Replacing the tokenizer with a CNN drops SR to $14.25\%$. The two
variants use the same data and $100$k-step training schedule, and the
gap favors explicit per-patch geometric features for metric lifting.
An inaccurate scale estimate can also make the learned lift worse than
fixed scaling.

\emph{Do lateral corrections help beyond per-waypoint radial scaling?}
A retrained radial-only variant predicts an independent radial scale
for each waypoint but no lateral residual. It reaches $27.62\%$ SR,
compared with $29.62\%$ for full PathLift, a $2.0$\,pp difference due
to the lateral correction.

\emph{Is the admissible-family supervision needed?}
Training the same network directly on $y^\star$, without $\Clift$,
reduces SR to $14.88\%$, again below the fixed-scale baseline.
The expert path $y^\star$ is a demonstration trajectory and need not
be a feasible metric realization of the current prior.

Combining all components yields the full $29.62\%$ SR.
Additional details are provided in App.~\ref{app:ablations}.

\begin{table}[!htbp]
\centering
\small
\setlength{\tabcolsep}{6pt}
\caption{
Comparison with local planners on the 800-trial NoMaD subset of
Tab.~\ref{tab:ablations} (bare NoMaD: $19.88\%$ SR).
DWA and MPPI take their metric goal from the fixed-scale lift, and
the composition variants take it from PathLift's lifted endpoint.
Runtime excludes the prior forward pass.
}
\vspace{0.5em}
\label{tab:planners}
\begin{tabular}{lrrr}
\toprule
\textbf{Method} & \textbf{SR (\%)} & \textbf{SPL (\%)} & \textbf{Stage (ms)} \\
\midrule
\rowcolor{gray!12}
PathLift                      & $29.62$          & $28.20$          & $\mathbf{5.6}$ \\
DWA                           & $26.75$          & $24.47$          & $13.8$ \\
MPPI                          & $34.12$          & $29.11$          & $21.3$ \\
\rowcolor{gray!12}
PathLift$\,\rightarrow\,$DWA  & $34.62$          & $31.89$          & $19.4$ \\
\rowcolor{gray!12}
PathLift$\,\rightarrow\,$MPPI & $\mathbf{38.00}$ & $\mathbf{34.57}$ & $26.9$ \\
\bottomrule
\end{tabular}
\end{table}

\subsection{Composition with Local Planners}
\label{sec:planners}

We compare PathLift with DWA and MPPI under the same protocol as
Sec.~\ref{sec:exp_ablation}. Both planners use a local costmap
constructed from the same depth frame provided to PathLift.
MPPI reaches $34.12\%$ SR, compared with $29.62\%$ for PathLift,
while DWA reaches $26.75\%$ (Tab.~\ref{tab:planners}).

In our setup, DWA and MPPI require a metric local goal that the
unitless prior cannot supply, so each runs downstream of a lift
rather than replacing it.
Using PathLift to provide this goal improves DWA from $26.75\%$ to
$34.62\%$ SR and MPPI from $34.12\%$ to $38.00\%$ SR.
PathLift$\rightarrow$MPPI achieves the highest SR and SPL
($38.00\%$ and $34.57\%$, respectively).
PathLift$\rightarrow$DWA also exceeds MPPI alone in both SR and SPL
while requiring less runtime.

\section{Conclusion and Limitations}
\label{sec:conclusion}

We presented PathLift, a depth-conditioned adapter for
visual-navigation policies. The adapter decouples three concerns:
the upstream policy (shape from a canonicalized prior), the depth
interface (sensor-agnostic camera-frame projection of RGB-D, stereo,
or LiDAR), and the lift method (per-waypoint magnitude and lateral
residual supervised by a structured admissible family of
intent-preserving trajectories). A single trained adapter therefore
composes with any pre-trained policy.

Against bare priors, PathLift raises closed-loop SR by
$35$--$78\%$ in simulation and $38$\,pp on a real wheeled robot
under stereo depth, with similar performance using projected LiDAR.
Against in-policy lifting trained per prior, our single adapter matches
NoMaD, trails FlowNav slightly, and runs over $10\times$ faster.
PathLift can also be used before a local planner, improving both
DWA and MPPI (Sec.~\ref{sec:planners}).

PathLift's success is bounded by the base prior's intent: when
the prior persistently points at obstacles or away from the goal,
the per-step lift cannot override a persistently misdirected prior. This is the cost of
off-the-shelf composability; PathLift's ceiling tracks the quality
of available pre-trained policies. The depth field of view is
matched to the prior's training FOV, trading outside-view
obstacle awareness for canonicalization consistency.
Real-robot evaluation is restricted to three indoor corridors.
The output parameterization (Eq.~\eqref{eq:output}) expresses
rotation only implicitly through per-waypoint magnitude and
lateral terms; an explicit rotational degree of freedom,
multi-frame depth fusion, higher-speed regimes, outdoor
deployment, and VLM coupling are natural next directions.

\acknowledgments{This work was supported by Hyundai Motor Company and Kia,
and by Institute of Information \& communications Technology Planning \&
Evaluation (IITP) grant funded by the Korea government(MSIT)
[NO.RS-2021-II211343, Artificial Intelligence Graduate School Program (Seoul
National University)].}

\clearpage
\bibliography{pathlift}

@inproceedings{shah2023gnm,
  title={{GNM}: A general navigation model to drive any robot},
  author={Shah, Dhruv and Sridhar, Ajay and Bhorkar, Arjun and Hirose, Noriaki and Levine, Sergey},
  booktitle={2023 IEEE International Conference on Robotics and Automation (ICRA)},
  pages={7226--7233},
  year={2023},
  organization={IEEE}
}

@inproceedings{shah2023vint,
  title={{ViNT}: A Foundation Model for Visual Navigation},
  author={Shah, Dhruv and Sridhar, Ajay and Dashora, Nitish and Stachowicz, Kyle and Black, Kevin and Hirose, Noriaki and Levine, Sergey},
  booktitle={Conference on Robot Learning},
  pages={711--733},
  year={2023},
  organization={PMLR}
}

@inproceedings{sridhar2024nomad,
  title={{NoMaD}: Goal masked diffusion policies for navigation and exploration},
  author={Sridhar, Ajay and Shah, Dhruv and Glossop, Catherine and Levine, Sergey},
  booktitle={2024 IEEE International Conference on Robotics and Automation (ICRA)},
  pages={63--70},
  year={2024},
  organization={IEEE}
}

@inproceedings{gode2025flownav,
  title={{FlowNav}: Combining flow matching and depth priors for efficient navigation},
  author={Gode, Samiran and Nayak, Abhijeet and Oliveira, D{\'e}bora NP and Krawez, Michael and Schmid, Cordelia and Burgard, Wolfram},
  booktitle={2025 IEEE/RSJ International Conference on Intelligent Robots and Systems (IROS)},
  pages={17762--17768},
  year={2025},
  organization={IEEE}
}

@inproceedings{ren2025navibridger,
  title={Prior does matter: Visual navigation via denoising diffusion bridge models},
  author={Ren, Hao and Zeng, Yiming and Bi, Zetong and Wan, Zhaoliang and Huang, Junlong and Cheng, Hui},
  booktitle={Proceedings of the IEEE/CVF Conference on Computer Vision and Pattern Recognition (CVPR)},
  pages={12100--12110},
  year={2025}
}

@article{zhang2025embodied,
  title={Embodied navigation foundation model},
  author={Zhang, Jiazhao and Li, Anqi and Qi, Yunpeng and Li, Minghan and Liu, Jiahang and Wang, Shaoan and Liu, Haoran and Zhou, Gengze and Wu, Yuze and Li, Xingxing and others},
  journal={arXiv preprint arXiv:2509.12129},
  year={2025}
}

@inproceedings{hirose2025omnivla,
  title={{OmniVLA}: An omni-modal vision-language-action model for robot navigation},
  author={Hirose, Noriaki and Glossop, Catherine and Shah, Dhruv and Levine, Sergey},
  booktitle={2026 IEEE International Conference on Robotics and Automation (ICRA)},
  pages={12651--12658},
  year={2026},
  organization={IEEE}
}

@inproceedings{bar2025nwm,
  title={Navigation world models},
  author={Bar, Amir and Zhou, Gaoyue and Tran, Danny and Darrell, Trevor and LeCun, Yann},
  booktitle={Proceedings of the IEEE/CVF Conference on Computer Vision and Pattern Recognition (CVPR)},
  pages={15791--15801},
  year={2025}
}

@article{guerrier2026can,
  title={Can Vision Foundation Models Navigate? Zero-Shot Real-World Evaluation and Lessons Learned},
  author={Guerrier, Maeva and Soma, Karthik and Pavlasek, Jana and Beltrame, Giovanni},
  journal={arXiv preprint arXiv:2603.25937},
  year={2026}
}

@inproceedings{nayak2025metricnet,
  title={{MetricNet}: Recovering Metric Scale in Generative Navigation Policies},
  author={Nayak, Abhijeet and Makowski, D{\'e}bora Oliveira and Gode, Samiran and Schmid, Cordelia and Burgard, Wolfram},
  booktitle={2026 IEEE International Conference on Robotics and Automation (ICRA)},
  pages={1863--1870},
  year={2026},
  organization={IEEE}
}

@inproceedings{zhang2026saferpath,
  title={{SaferPath}: Hierarchical Visual Navigation with Learned Guidance and Safety-Constrained Control},
  author={Zhang, Lingjie and Jiang, Zeyu and Chen, Changhao},
  booktitle={2026 IEEE International Conference on Robotics and Automation (ICRA)},
  pages={18259--18266},
  year={2026},
  organization={IEEE}
}

@inproceedings{cai2025navdp,
  title={{NavDP}: Learning sim-to-real navigation diffusion policy with privileged information guidance},
  author={Cai, Wenzhe and Peng, Jiaqi and Yang, Yuqiang and Zhang, Yujian and Wei, Meng and Wang, Hanqing and Chen, Yilun and Wang, Tai and Pang, Jiangmiao},
  booktitle={2026 IEEE International Conference on Robotics and Automation (ICRA)},
  pages={5233--5240},
  year={2026},
  organization={IEEE}
}

@inproceedings{ratliff2009chomp,
  title={{CHOMP}: Gradient optimization techniques for efficient motion planning},
  author={Ratliff, Nathan and Zucker, Matt and Bagnell, J Andrew and Srinivasa, Siddhartha},
  booktitle={2009 IEEE International Conference on Robotics and Automation (ICRA)},
  pages={489--494},
  year={2009},
  organization={IEEE}
}

@inproceedings{rosmann2012trajectory,
  title={Trajectory modification considering dynamic constraints of autonomous robots},
  author={R{\"o}smann, Christoph and Feiten, Wendelin and W{\"o}sch, Thomas and Hoffmann, Frank and Bertram, Torsten},
  booktitle={ROBOTIK 2012; 7th German Conference on Robotics},
  pages={1--6},
  year={2012},
  organization={VDE}
}

@article{shi2022motion,
  title={Motion transformer with global intention localization and local movement refinement},
  author={Shi, Shaoshuai and Jiang, Li and Dai, Dengxin and Schiele, Bernt},
  journal={Advances in Neural Information Processing Systems},
  volume={35},
  pages={6531--6543},
  year={2022}
}

@inproceedings{usenko2018double,
  title={The double sphere camera model},
  author={Usenko, Vladyslav and Demmel, Nikolaus and Cremers, Daniel},
  booktitle={2018 International Conference on 3D Vision (3DV)},
  pages={552--560},
  year={2018},
  organization={IEEE}
}

@article{sitzmann2021light,
  title={Light field networks: Neural scene representations with single-evaluation rendering},
  author={Sitzmann, Vincent and Rezchikov, Semon and Freeman, Bill and Tenenbaum, Josh and Durand, Fredo},
  journal={Advances in Neural Information Processing Systems},
  volume={34},
  pages={19313--19325},
  year={2021}
}

@article{yang2024depth,
  title={{Depth Anything V2}},
  author={Yang, Lihe and Kang, Bingyi and Huang, Zilong and Zhao, Zhen and Xu, Xiaogang and Feng, Jiashi and Zhao, Hengshuang},
  journal={Advances in Neural Information Processing Systems},
  volume={37},
  pages={21875--21911},
  year={2024}
}

@inproceedings{wen2025fast,
  title={{Fast-FoundationStereo}: Real-Time Zero-Shot Stereo Matching},
  author={Wen, Bowen and Dewan, Shaurya and Birchfield, Stan},
  booktitle={Proceedings of the IEEE/CVF Conference on Computer Vision and Pattern Recognition (CVPR)},
  year={2026}
}

@inproceedings{savva2019habitat,
  title={Habitat: A platform for embodied {AI} research},
  author={Savva, Manolis and Kadian, Abhishek and Maksymets, Oleksandr and Zhao, Yili and Wijmans, Erik and Jain, Bhavana and Straub, Julian and Liu, Jia and Koltun, Vladlen and Malik, Jitendra and Parikh, Devi and Batra, Dhruv},
  booktitle={Proceedings of the IEEE/CVF International Conference on Computer Vision (ICCV)},
  pages={9339--9347},
  year={2019}
}

@inproceedings{ramakrishnan2021hm3d,
  title={{Habitat-Matterport 3D Dataset (HM3D)}: 1000 Large-scale {3D} Environments for Embodied {AI}},
  author={Santhosh Kumar Ramakrishnan and Aaron Gokaslan and Erik Wijmans and Oleksandr Maksymets and Alexander Clegg and John M Turner and Eric Undersander and Wojciech Galuba and Andrew Westbury and Angel X Chang and Manolis Savva and Yili Zhao and Dhruv Batra},
  booktitle={Thirty-fifth Conference on Neural Information Processing Systems Datasets and Benchmarks Track (Round 2)},
  year={2021},
  url={https://openreview.net/forum?id=-v4OuqNs5P}
}

@article{zhang2024toward,
  title={Toward robust robot {3-D} perception in urban environments: {The UT Campus Object Dataset}},
  author={Zhang, Arthur and Eranki, Chaitanya and Zhang, Christina and Park, Ji-Hwan and Hong, Raymond and Kalyani, Pranav and Kalyanaraman, Lochana and Gamare, Arsh and Bagad, Arnav and Esteva, Maria and Biswas, Joydeep},
  journal={IEEE Transactions on Robotics},
  volume={40},
  pages={3322--3340},
  year={2024},
  publisher={IEEE}
}

@inproceedings{shah2022rapid,
  title={Rapid Exploration for Open-World Navigation with Latent Goal Models},
  author={Shah, Dhruv and Eysenbach, Benjamin and Rhinehart, Nicholas and Levine, Sergey},
  booktitle={Conference on Robot Learning},
  pages={674--684},
  year={2022},
  organization={PMLR}
}

@inproceedings{shi2020we,
  title={Are we ready for service robots? {The OpenLORIS-Scene} datasets for lifelong {SLAM}},
  author={Shi, Xuesong and Li, Dongjiang and Zhao, Pengpeng and Tian, Qinbin and Tian, Yuxin and Long, Qiwei and Zhu, Chunhao and Song, Jingwei and Qiao, Fei and Song, Le and others},
  booktitle={2020 IEEE International Conference on Robotics and Automation (ICRA)},
  pages={3139--3145},
  year={2020},
  organization={IEEE}
}

@article{hirose2023sacson,
  title={{SACSoN}: Scalable autonomous control for social navigation},
  author={Hirose, Noriaki and Shah, Dhruv and Sridhar, Ajay and Levine, Sergey},
  journal={IEEE Robotics and Automation Letters},
  volume={9},
  number={1},
  pages={49--56},
  year={2024},
  publisher={IEEE}
}

@article{anderson2018evaluation,
  title={On evaluation of embodied navigation agents},
  author={Anderson, Peter and Chang, Angel and Chaplot, Devendra Singh and Dosovitskiy, Alexey and Gupta, Saurabh and Koltun, Vladlen and Kosecka, Jana and Malik, Jitendra and Mottaghi, Roozbeh and Savva, Manolis and Zamir, Amir R.},
  journal={arXiv preprint arXiv:1807.06757},
  year={2018}
}

@inproceedings{liu2025citywalker,
  title={{CityWalker}: Learning embodied urban navigation from web-scale videos},
  author={Liu, Xinhao and Li, Jintong and Jiang, Yicheng and Sujay, Niranjan and Yang, Zhicheng and Zhang, Juexiao and Abanes, John and Zhang, Jing and Feng, Chen},
  booktitle={Proceedings of the IEEE/CVF Conference on Computer Vision and Pattern Recognition (CVPR)},
  pages={6875--6885},
  year={2025}
}

@inproceedings{schonberger2016structure,
  title={Structure-from-motion revisited},
  author={Sch{\"o}nberger, Johannes L. and Frahm, Jan-Michael},
  booktitle={Proceedings of the IEEE Conference on Computer Vision and Pattern Recognition (CVPR)},
  pages={4104--4113},
  year={2016}
}

@article{teed2021droid,
  title={{DROID-SLAM}: Deep visual {SLAM} for monocular, stereo, and {RGB-D} cameras},
  author={Teed, Zachary and Deng, Jia},
  journal={Advances in Neural Information Processing Systems},
  volume={34},
  pages={16558--16569},
  year={2021}
}

@article{teed2023deep,
  title={Deep patch visual odometry},
  author={Teed, Zachary and Lipson, Lahav and Deng, Jia},
  journal={Advances in Neural Information Processing Systems},
  volume={36},
  pages={39033--39051},
  year={2023}
}

@inproceedings{leroy2024grounding,
  title={Grounding image matching in {3D} with {MASt3R}},
  author={Leroy, Vincent and Cabon, Yohann and Revaud, J{\'e}r{\^o}me},
  booktitle={European Conference on Computer Vision (ECCV)},
  pages={71--91},
  year={2024},
  organization={Springer}
}

@inproceedings{wang2025vggt,
  title={{VGGT}: Visual geometry grounded transformer},
  author={Wang, Jianyuan and Chen, Minghao and Karaev, Nikita and Vedaldi, Andrea and Rupprecht, Christian and Novotny, David},
  booktitle={Proceedings of the IEEE/CVF Conference on Computer Vision and Pattern Recognition (CVPR)},
  pages={5294--5306},
  year={2025}
}

@inproceedings{lee2015deeply,
  title={Deeply-supervised nets},
  author={Lee, Chen-Yu and Xie, Saining and Gallagher, Patrick and Zhang, Zhengyou and Tu, Zhuowen},
  booktitle={International Conference on Artificial Intelligence and Statistics (AISTATS)},
  pages={562--570},
  year={2015},
  organization={PMLR}
}

@article{shen2025effonav,
  title={{EffoNAV}: An Effective Foundation-Model-Based Visual Navigation Approach in Challenging Environment},
  author={Shen, Wangtian and Gu, Pengfei and Qin, Haijian and Meng, Ziyang},
  journal={IEEE Robotics and Automation Letters},
  volume={10},
  number={7},
  pages={6824--6831},
  year={2025},
  publisher={IEEE}
}

@article{karnan2022socially,
  title={Socially compliant navigation dataset ({SCAND}): A large-scale dataset of demonstrations for social navigation},
  author={Karnan, Haresh and Nair, Anirudh and Xiao, Xuesu and Warnell, Garrett and Pirk, S{\"o}ren and Toshev, Alexander and Hart, Justin and Biswas, Joydeep and Stone, Peter},
  journal={IEEE Robotics and Automation Letters},
  volume={7},
  number={4},
  pages={11807--11814},
  year={2022},
  publisher={IEEE}
}

@inproceedings{triest2022tartandrive,
  title={{TartanDrive}: A large-scale dataset for learning off-road dynamics models},
  author={Triest, Samuel and Sivaprakasam, Matthew and Wang, Sean J and Wang, Wenshan and Johnson, Aaron M and Scherer, Sebastian},
  booktitle={2022 International Conference on Robotics and Automation (ICRA)},
  pages={2546--2552},
  year={2022},
  organization={IEEE}
}

@article{schulman2014motion,
  title={Motion planning with sequential convex optimization and convex collision checking},
  author={Schulman, John and Duan, Yan and Ho, Jonathan and Lee, Alex and Awwal, Ibrahim and Bradlow, Henry and Pan, Jia and Patil, Sachin and Goldberg, Ken and Abbeel, Pieter},
  journal={The International Journal of Robotics Research},
  volume={33},
  number={9},
  pages={1251--1270},
  year={2014},
  publisher={Sage Publications Sage UK: London, England}
}

\clearpage
\appendix

\section{Candidate sub-families}
\label{app:tiers}

Let $\arc \in[0,1]^H$ denote per-waypoint cumulative arc fraction on
$\phat$ and $\mathbf{n}(\phat)\in\mathbb{R}^{H\times 2}$ the per-waypoint
unit normals. Let $\mathcal{R}$ be the radius grid
$\{0.40,0.49,\ldots,3.00\}$ ($19$ values mixing linear-dense near $1.0$
with log-spaced extension to a per-source $R_{\max}$: $14.0$\,m (CODa~\citep{zhang2024toward}
outdoor), $10.0$\,m (Habitat~\citep{savva2019habitat} indoor), and $6.0$\,m (OpenLORIS~\citep{shi2020we} indoor),
calibrated to each domain's path-length distribution). Tiers that
compose a shape parameter with radius subsample $\mathcal{R}$ to keep
each sub-family's count bounded.

Table~\ref{tab:tiers} lists the ten sub-families, their
shape-parameter spaces, and per-family candidate counts. The total is
$|\Clift|=679$.

\begin{table}[h]
\centering\small
\renewcommand{\arraystretch}{1.15}
\caption{Candidate sub-family parameterizations. $\mathbf{R}_z(\theta)$ is
the $2$-D rotation around the body origin; $f(\arc)$ are radial
schedules; $b$ is the B\'ezier lateral envelope
$b(\arc;\alpha_m,\alpha_e)=2\alpha_m\,\arc(1-\arc)+\alpha_e\,\arc^2$;
$\rho$ and $e$ are cubic smoothstep envelopes that anchor the lateral
component at the body origin. $T_7$ is the rescue sub-family with an enlarged lateral envelope; the
admissibility filter prunes it whenever a standard-envelope candidate
is admissible.}
\vspace{0.5em}
\label{tab:tiers}
\begin{tabularx}{\linewidth}{@{}ll>{\raggedright\arraybackslash}Xc@{}}
\toprule
Sub-family & $c_t(\phat; R, \theta)$ & Shape parameter $\theta$ & Count \\
\midrule
$T_0$ identity
& $\phat$
& --- & $1$ \\
$T_1$ radial
& $R\,\phat$
& --- & $|\mathcal{R}| = 19$ \\
$T_2$ smooth radial
& $R\,f(\arc)\,\phat$
& $f\in\{\,\mathrm{slow,\ fast,\ sat_{0.7},\ sat_{0.5}}\,\}$ \par
  \hspace{1.5em}or quadratic basis with $(a_1,a_2)\!\in\!\{-0.4,0,0.4\}^2$
& $40+45=85$ \\
$T_3$ progress
& $R\,\phat\!\circ\!w(\arc)$
& truncation $\tau\!\in\!\{0.4,0.55,0.7,0.85,1.0\}$ \par
  \hspace{1.5em}or gamma $\gamma\!\in\!\{0.5,0.75,1,1.5,2\}$
& $50+50=100$ \\
$T_4$ rotation
& $R\,\mathbf{R}_z(\theta)\,\phat$
& $\theta\!\in\!\pm\{1.5,3,5,8,12,20,30\}^\circ$
& $14\times 7 = 98$ \\
$T_5$ rotation $\times$ smooth radial
& $R\,f(\arc)\,\mathbf{R}_z(\theta)\,\phat$
& $\theta\!\in\!\{-30,-12,-5,-1.5,3,8,20\}^\circ$, $f\!\in\!\{\mathrm{slow,fast}\}$
& $7\times 2\times 5 = 70$ \\
$T_6$ B\'ezier lateral
& $R\,\phat + b(\arc;\alpha_m,\alpha_e)\mathbf{n}(\phat)$
& $(\alpha_m,\alpha_e)\!\in\!\{-0.3,\!{-0.15},0,0.15,0.3\}\!\times$ \par
  \hspace{1.5em}$\{-0.4,\!{-0.2},0,0.2,0.4\}$ minus identity
& $24\times 7 = 168$ \\
$T_{6b}$ constant lateral
& $R\,\phat + d_\mathrm{lat}\,\rho(\arc)\,\mathbf{n}(\phat)$
& $d_\mathrm{lat}\!\in\!\{\pm 0.3,\pm 0.2,\pm 0.1\}$, $\rho$ smoothstep
& $6\times 7 = 42$ \\
$T_7$ rescue
& $R\,\mathbf{R}_z(\theta)\,\phat + b(\arc;0,\alpha_e)\mathbf{n}(\phat)$
& $\theta\!\in\!\pm\{60,75,90\}^\circ$, $\alpha_e\!\in\!\{-0.4,0,0.4\}$, $R\!\in\!\{1,2,4\}$
& $6\times 3\times 3 = 54$ \\
$T_8$ endpoint deflection
& $R\,\phat + \alpha\,e(\arc; s_0\!=\!0.7)\mathbf{n}(\phat)$
& $\alpha\!\in\!\{\pm 0.4,\pm 0.25,\pm 0.12\}$, $e$ smoothstep beyond $s_0$
& $6\times 7 = 42$ \\
\midrule
& & \textbf{Total} & $\mathbf{679}$ \\
\bottomrule
\end{tabularx}
\end{table}

\begin{table}[h]
\centering\footnotesize
\setlength{\tabcolsep}{6pt}
\caption{Admissibility pass rate (\%) per sub-family, averaged
within each source. Standard sub-families fall in a uniform
$35$--$56\%$ indoor range; T$_7$ rescue passes $0.1$--$2.3\%$,
consistent with its role as a fallback pruned whenever a
standard-envelope candidate is admissible.}
\vspace{0.5em}
\label{tab:tiers_admit}
\begin{tabular}{l rrrr}
\toprule
Sub-family & Habitat pin. & Habitat fish. & CODa & OpenLORIS \\
\midrule
T$_0$ identity            & 48.5 & 55.6 & 95.4 & 60.5 \\
T$_1$ radial              & 44.0 & 49.4 & 94.0 & 52.4 \\
T$_2$ smooth radial       & 47.6 & 53.0 & 94.7 & 55.6 \\
T$_3$ progress            & 43.0 & 47.4 & 88.7 & 50.9 \\
T$_4$ rotation            & 40.6 & 44.9 & 81.9 & 46.1 \\
T$_5$ rotation $\times$ radial & 45.1 & 49.9 & 84.9 & 50.1 \\
T$_6$ B\'ezier lateral    & 34.7 & 39.6 & 83.2 & 37.6 \\
T$_{6b}$ constant lateral & 36.3 & 43.6 & 91.3 & 41.2 \\
T$_7$ rescue              & 0.2  & 0.4  & 2.3  & 0.1  \\
T$_8$ endpoint deflection & 40.6 & 46.0 & 91.9 & 46.7 \\
\bottomrule
\end{tabular}
\end{table}

The admissibility filter (Eq.~\eqref{eq:adm}) shapes the
supervision target by reducing the $679$-candidate family to a
non-trivial depth-feasible subset. Standard sub-families pass
$35$--$56\%$ of the time indoor (Tab.~\ref{tab:tiers_admit}),
demonstrating the filter's selectivity, and rise to $82$--$95\%$
on outdoor CODa where admissibility constraints are largely
unbinding. The same family and thresholds serve all four sources
without per-source tuning.

\section{Admissibility threshold sensitivity}
\label{app:sensitivity}
Recall from \S\ref{sec:supervision} that 
$\mathrm{tube}(c) = \tfrac{1}{H}\sum_k \|c_k - \alpha_c\hat{p}_k\|$ 
with $\alpha_c = \langle c, \hat{p}\rangle / \|\hat{p}\|^2$ 
measures the average residual after best-fit radial alignment of 
the prior with the candidate.
The admissibility filter (Eq.~\ref{eq:adm}) uses four predicate 
thresholds, but they play structurally different roles. 
The clearance ($\tau_\mathrm{clr}=0.30$\,m) and penetration 
($\tau_\mathrm{pen}=0.20$\,m) thresholds derive from the robot's 
collision tolerance, and the observation floor 
($\tau_\mathrm{obs}=0.90$) is tied to the camera FOV. The 
intent-preservation tube ($\tau_\mathrm{tube}=0.30$\,m) has no 
such physical anchor. It is a free parameter that trades prior 
fidelity against candidate breadth. We therefore focus the
sensitivity analysis on $\tau_\mathrm{tube}$, sweeping it from
$0.15$ to $0.50$\,m around the paper value and reporting the
$c^\star$ selection change rate against the
$\tau_\mathrm{tube}=0.30$\,m reference together with
$\mathrm{ADE}(c^\star,y^\star)$ (Tab.~\ref{tab:sens_combined}).

\begin{table}[h]
\centering
\footnotesize
\setlength{\tabcolsep}{4pt}
\caption{
Sensitivity to the intent-preservation threshold
$\tau_\mathrm{tube}$, with the other admissibility thresholds fixed.
The left block reports the fraction of samples whose selected
$c^\star$ changes by more than $0.05$\,m relative to
$\tau_\mathrm{tube}=0.30$.
The right block reports $\mathrm{ADE}(c^\star,y^\star)$.
OpenLORIS is omitted because its expert paths often exceed the
per-source candidate range $R_{\max}$ (App.~\ref{app:tiers}),
making the selection insensitive to this threshold.
}
\vspace{0.5em}
\label{tab:sens_combined}
\begin{tabular}{l rrrrrr rrr}
\toprule
 & \multicolumn{6}{c}{$c^\star$ change rate (\%)}
 & \multicolumn{3}{c}{$\mathrm{ADE}(c^\star,y^\star)$ (m)} \\
\cmidrule(lr){2-7}\cmidrule(lr){8-10}
Source & $0.15$ & $0.20$ & $0.25$ & $0.35$ & $0.40$ & $0.50$
 & $0.15$ & $0.30$ & $0.50$ \\
\midrule
Habitat pinhole & 48.5 & 31.0 & 10.9 & 6.9 & 16.9 & 18.6 & 0.616 & 0.573 & 0.553 \\
Habitat fisheye & 43.1 & 29.0 & 13.9 & 6.6 & 15.1 & 17.2 & 0.644 & 0.611 & 0.595 \\
CODa            & 62.4 & 29.4 & 17.1 & 36.0 & 41.9 & 56.4 & 2.947 & 2.776 & 2.593 \\
\bottomrule
\end{tabular}
\end{table}

Relaxing $\tau_\mathrm{tube}$ enlarges the admissible set, so
$\mathrm{ADE}(c^\star,y^\star)$ can only decrease, as the table shows.
For Habitat pinhole, increasing the threshold from $0.30$ to
$0.50$\,m improves ADE by only $3.5\%$, while tightening it to
$0.15$\,m increases ADE by $7.5\%$ and changes the selected target
for $48.5\%$ of samples.
We therefore use $0.30$\,m as a practical balance between preserving
the prior and retaining a sufficiently broad candidate set.
CODa is more sensitive to larger thresholds because its open outdoor
scenes leave more candidates admissible.

\section{Architecture, loss weights, and training schedule}
\label{app:loss}

\paragraph{Lifting horizon.} The candidate family $\Clift$ and the
supervision target $c^\star$ are generated at a fixed maximum
horizon $H{=}24$, a $3\times$ upsampling of the native $8$-waypoint
cumsum output common to most visual-navigation policies (NoMaD, ViNT,
GNM); this gives finer metric resolution along the path while keeping
anchor-decoder computation linear in $H$. The network is exercised at
variable horizons $H_\text{batch}\!\in\!\{5,8,12,16,24\}$ per
minibatch during training by truncating the prior, supervision target,
and waypoint tokens together to $H_\text{batch}$; output length
matches the truncated input. At deployment, PathLift can be queried
at any $H_\text{deploy}\!\le\!24$.

\paragraph{Network sizes.}
$224\!\times\!224$ depth input, patch size $14$. Hidden width
$d = 128$, $4$ attention heads; total parameters $4.3$\,M.
Transformer blocks use pre-LayerNorm with GELU and FFN expansion
ratio $4$; tokens carry no per-position positional encoding, since
geometric position is encoded directly in the token features
(Pl\"ucker ray for patches, canonical position for waypoints).

\paragraph{Loss weights.} Table~\ref{tab:lambdas} lists all training
loss weights. The principal term is $\mathcal{L}_{\mathrm{lift}}$, the
squared error to the supervision target; auxiliary terms enforce
admissibility (clearance hinge), prior coupling (equivariance, tangent
alignment), trajectory smoothness, and per-block deep supervision.

\begin{table}[h]
\centering\small
\caption{Loss-term weights. All values are fixed across experiments
and visual-navigation policies.}
\vspace{0.5em}
\label{tab:lambdas}
\begin{tabular}{@{}lcl@{}}
\toprule
Term & Weight & Role \\
\midrule
$\mathcal{L}_{\mathrm{lift}}$       & $1$    & Squared error to supervision target $c^\star$ \\
$\mathcal{L}_{\mathrm{clr}}$        & $0.2$  & Clearance hinge against $\Omega$ \\
$\mathcal{L}_{\mathrm{equiv}}$      & $0.1$  & Magnitude equivariance regularizer \\
$\mathcal{L}_{\mathrm{tan{-}prior}}$& $0.5$  & Tangent alignment to $\phat$ \\
$\mathcal{L}_{r\text{-aux}}$        & $0.1$  & Auxiliary log-magnitude regression \\
$\mathcal{L}_{\mathrm{curv}}$       & $0.05$ & Second-difference smoothness \\
$\mathcal{L}_{\mathrm{deep\text{-}sup}}$ & $0.25$ & Per-block decoder supervision \\
\bottomrule
\end{tabular}
\end{table}

\paragraph{Optimization.} AdamW with weight decay $10^{-4}$, cosine
annealing on a learning rate of $4.24\!\times\!10^{-4}$
($=3\!\times\!10^{-4}\!\cdot\!\sqrt{2}$ for batch size $64$). Training
runs for $100$k steps with batch size $64$, without mixed precision.
Horizons $H_\mathrm{batch}\!\in\!\{5,8,12,16,24\}$ are sampled
uniformly per minibatch by truncating $\phat$, the supervision target
$c^\star$, and all auxiliary terms; this matches variable-horizon controller usage
and regularizes against endpoint-specific decoder boundary effects.
Training runs on a single NVIDIA RTX PRO 6000.

\section{Simulation Experiment Details}
\label{app:sim_setup}

\subsection{Robot, Sensors, and Control}
\label{app:sim_robot}
The agent has a radius of $0.3$\,m, with camera mount height
randomized between $0.25$\,m and $1.0$\,m per trial. RGB input is
captured by a $640\!\times\!480$ fisheye camera ($\text{HFOV}=
170^{\circ}$), and depth input by a $224\!\times\!224$ pinhole camera
($\text{HFOV}=90^{\circ}$). Depth values beyond a fixed range are
masked as NaN to replicate the noise and range limits of a real
RGB-D sensor. Control is performed by teleportation, where a
low-level controller tracks the metric waypoint sequence output by
the model at each inference step.

\subsection{MetricNav with Ground-Truth Depth}
\label{app:metricnav_star}
MetricNav~\citep{nayak2025metricnet} recovers scale through MetricNet
and constructs a TSDF map from
DepthAnything~\citep{yang2024depth} to provide collision-cost
and goal-cost guidance during the denoising loop of the base prior.
Its intended deployment therefore uses DepthAnything as the depth
source. In our benchmark, however, PathLift consumes ground-truth
depth, so a direct comparison would conflate adapter design with
depth-source quality. For a fair comparison on the same input
modality, we additionally report MetricNav$^{*}$, in which the TSDF
is built from the same ground-truth depth used by PathLift; all other
components of MetricNav (scalar prediction, sampling, cost
computation, denoising-step guidance) are left untouched.

\subsection{Finetuning Protocol}
\label{app:finetuning}
The official pretrained checkpoints of each prior are trained on
real-world mobile-robot datasets (e.g.,
SACSoN~\citep{hirose2023sacson}, RECON~\citep{shah2022rapid}, SCAND~\citep{karnan2022socially}, TartanDrive~\citep{triest2022tartandrive})
whose embodiments, camera geometry, and scene statistics differ from
indoor HM3D~\citep{ramakrishnan2021hm3d} scenes, causing domain shift. We collected
approximately $20{,}000$ trajectories from HM3D scenes disjoint from
our benchmark set within the same Habitat-sim setup and converted
them into ego-frame waypoints and RGB sequences to finetune all four
priors under identical conditions. Preprocessing involves (a)
aligning Habitat $XZ$-plane coordinates to the robot's local
$XY$-frame, (b) applying a sliding window with an $8$-step prediction
horizon, and (c) standardizing the normalized cumulative
displacement per step. All priors were trained on a single NVIDIA
RTX 4090 for $25$ epochs with a context size of $3$, a batch size of
$256$, and a learning rate of $5\!\times\!10^{-5}$.

\section{Real-Robot Deployment Details}
\label{app:real_deployment}

This appendix expands the real-robot deployment of
Sec.~\ref{sec:real_robot}: the platform specification
(Tab.~\ref{tab:real_platform}) and the depth-pathway pipeline.

\subsection{Platform and Sensing}
\label{app:real_platform}

\begin{table}[h]
\centering\footnotesize
\setlength{\tabcolsep}{6pt}
\renewcommand{\arraystretch}{1.1}
\caption{Real-robot deployment platform.}
\vspace{0.5em}
\label{tab:real_platform}
\begin{tabular}{@{}ll@{}}
\toprule
\textbf{Component} & \textbf{Specification} \\
\midrule
Base                  & indoor wheeled differential-drive mobile robot \\
Forward RGB camera    & stereo pair, left lens HFOV $= 90^{\circ}$,
                        $640\!\times\!480$ per channel \\
LiDAR                 & MID-360; top-mounted, downward-facing scan plane \\
Stereo depth model    & Fast-FoundationStereo~\citep{wen2025fast} \\
Onboard compute       & NVIDIA Jetson AGX Orin (64\,GB) \\
Controller            & pure-pursuit tracker at $4$\,Hz \\
\bottomrule
\end{tabular}
\end{table}

\paragraph{Depth pathway.}
Both depth sources feed the same PathLift checkpoint without
sensor-specific tuning. Stereo depth (SD) is produced by
Fast-FoundationStereo~\citep{wen2025fast} on each forward stereo
frame at the controller rate. The LiDAR scan is re-projected through
the camera projection $\proj$ (Sec.~\ref{sec:arch}) into the
forward camera frame, so PathLift consumes a depth image with the
same intrinsics and extrinsics regardless of which sensor produced
it. This shared interface is what makes the two sources
interchangeable in the results of
Sec.~\ref{sec:real_imagegoal}.

\section{Ablation Details}
\label{app:ablations}
\paragraph{Protocol.}
All ablations remove one component at a time from the canonical
PathLift configuration, keeping the rest fixed. We use the NoMaD prior
on the $800$-trial subset ($20$ HM3D validation scenes $\times$ $20$
trajectories $\times$ $2$ repetitions); the bare NoMaD baseline
reaches $19.88\%$ SR on this pool (Tab.~\ref{tab:ablations_full}).

\begin{table}[!htbp]
\centering\small
\setlength{\tabcolsep}{6pt}
\caption{Full ablation results on the $800$-trial NoMaD subset.
Each block removes one component from PathLift while keeping the
rest fixed; setup details are in the corresponding paragraphs.}
\vspace{0.5em}
\label{tab:ablations_full}
\begin{tabular}{lcc}
\toprule
\textbf{Variant} & \textbf{Enc.~params} & \textbf{SR (\%)} \\
\midrule
\rowcolor{gray!12}
PathLift (Pl\"ucker, per-wp)                          & $2.3$\,k  & $\mathbf{29.62}$ \\
\midrule
\multicolumn{3}{l}{\emph{Depth ablation}} \\
$-$\,depth input                                      & ---       & $6.50$  \\
\midrule
\multicolumn{3}{l}{\emph{Encoder ablation (input tokenization)}} \\
$-$\,Pl\"ucker tokenization (CNN encoder)             & $37$\,k   & $14.25$ \\
\midrule
\multicolumn{3}{l}{\emph{Output-head ablation}} \\
$-$\,Radial-only (per-waypoint)        & ---       & $27.62$ \\
\midrule
\multicolumn{3}{l}{\emph{Supervision ablation}}\\
$-$\,$\Clift$ supervision (direct $y^\star$ regression) & --- & $14.88$ \\
\midrule
Bare NoMaD (fixed-scale lift)                & ---       & $19.88$ \\
\bottomrule
\end{tabular}
\end{table}

\paragraph{Depth input.}
We ablate the depth signal by overwriting every pixel of the input
depth map with $-1.0$\,m, the value used to mark invalid pixels
during training. The shape contract of the adapter is preserved,
and the Pl\"ucker rays, camera intrinsics, prior trajectory, and
scene encoder weights are unchanged. We use this in-distribution
value rather than an arbitrary constant (e.g. $5$\,m), which
would instead test an ill-defined failure mode where the model
would confidently infer a flat wall everywhere. Under this
protocol SR drops to $6.50\%$, $13.38$\,pp below the bare-prior
baseline (Tab.~\ref{tab:ablations_full}).

\paragraph{Partial-invalid sweep.}
The depth ablation above invalidates the entire input. We additionally
vary the fraction of invalid pixels in open loop
(Tab.~\ref{tab:invalid_sweep}). ADE changes by less than
$0.001$\,m up to $95\%$ invalid pixels and by $0.007$\,m at
$99\%$. A clear drop appears only when all pixels are invalid, where
the prediction moves toward a pure radial lift. Even with $99\%$
invalid pixels, the remaining valid depth samples are sufficient to
recover metric scale from the patch statistics.

\begin{table}[!htbp]
\centering
\small
\setlength{\tabcolsep}{4.5pt}
\caption{
Partial-invalid depth sweep on $800$ Habitat-pinhole samples with
the NoMaD prior. Random pixels are replaced by the training-time
invalid value $-1$, using nested masks across invalid fractions.
}
\vspace{0.5em}
\label{tab:invalid_sweep}
\begin{tabular}{l rrrrrrrrr}
\toprule
Invalid (\%) & $0$ & $10$ & $25$ & $50$ & $75$ & $90$ & $95$ & $99$ & $100$ \\
\midrule
ADE (m)      & $0.232$ & $0.232$ & $0.232$ & $0.232$ & $0.232$ & $0.233$ & $0.233$ & $0.239$ & $0.413$ \\
tube (m)     & $0.079$ & $0.079$ & $0.079$ & $0.079$ & $0.079$ & $0.079$ & $0.079$ & $0.072$ & $0.030$ \\
$\Cadm$ (\%) & $68.5$ & $68.8$ & $68.9$ & $68.9$ & $69.0$ & $68.6$ & $69.1$ & $68.4$ & $64.4$ \\
\bottomrule
\end{tabular}
\end{table}

\paragraph{Encoder.}
The linear Pl\"ucker patch adapter has $2.3$\,k parameters. The
CNN variant has $37$\,k parameters ($\sim\!16\times$ more) and
processes the same $224\!\times\!224$ depth map without explicit
ray or pose channels. The CNN reaches $14.25\%$ SR, compared with
$29.62\%$ for the Pl\"ucker variant and $19.88\%$ for the bare
prior (Tab.~\ref{tab:ablations_full}).

\paragraph{Output head.}
We retrain a radial-only variant that keeps the per-waypoint
magnitude head and removes the lateral residual head. It predicts
an independent radial scale at each waypoint without lateral
correction. This variant reaches $27.62\%$ SR, compared with
$29.62\%$ for full PathLift (Tab.~\ref{tab:ablations_full}).
The remaining $2.0$\,pp gap corresponds to the addition of lateral
corrections in the full model.

\paragraph{Supervision.}
Removing $\Clift$ and training the same architecture directly on the
arc-length-resampled expert path $y^\star$ reduces SR to $14.88\%$,
below the fixed-scale baseline. The expert path is a demonstration
and need not be a feasible metric realization of the current prior.
The admissible family instead restricts the target to depth-feasible
realizations of the prior, with $y^\star$ used to select among them.

\end{document}